**Single-shot reconstruction of three-dimensional morphology of biological cells in digital holographic microscopy using a physics-driven neural network**

Jihwan Kim[1], Youngdo Kim[1], Hyo Seung Lee[1], Eunseok Seo[2], Sang Joon Lee[1*]

[1]Department of Mechanical Engineering, Pohang University of Science and Technology, Pohang, 37673, Republic of Korea

[2]Department of Mechanical Engineering, Sogang University, Seoul, 04107, Republic of Korea

*Corresponding author

E-mail: sjlee@postech.ac.kr

Phone: +82-54-279-2169

Fax: +82-54-279-3199

**Abstract**

Recent advances in deep learning-based image reconstruction techniques have led to significant progress in phase retrieval using digital in-line holographic microscopy (DIHM). However, existing deep learning-based phase retrieval methods have technical limitations in generalization performance and three-dimensional (3D) morphology reconstruction from a single-shot hologram of biological cells. In this study, we propose a novel deep learning model, named MorpHoloNet, for single-shot reconstruction of 3D morphology by integrating physics-driven and coordinate-based neural networks. By simulating the optical diffraction of coherent light through a 3D phase shift distribution, the proposed MorpHoloNet is optimized by minimizing the loss between the simulated and input holograms on the sensor plane. Compared to existing DIHM methods that face challenges with twin image and phase retrieval problems, MorpHoloNet enables direct reconstruction of 3D complex light field and 3D morphology of a test sample from its single-shot hologram without requiring multiple phase-shifted holograms or angle scanning. The performance of the proposed MorpHoloNet is validated by reconstructing 3D morphologies and refractive index distributions from synthetic holograms of ellipsoids and experimental holograms of biological cells. The proposed deep learning model is utilized to reconstruct spatiotemporal variations in 3D translational and rotational behaviors and morphological deformations of biological cells from consecutive single-shot holograms captured using DIHM. MorpHoloNet would pave the way for advancing label-free, real-time 3D imaging and dynamic analysis of biological cells under various cellular microenvironments in biomedical and engineering fields.

# 1. Introduction

Digital holographic microscopy (DHM) is a label-free three-dimensional (3D) imaging technique which has been widely utilized in various biomedical and engineering fields[1-3]. Holographic patterns generated by optical interferences between object and reference beams from a coherent light source are captured by a digital imaging device, such as a charge-coupled device camera or a complementary metal-oxide-semiconductor (CMOS) camera. The complex light fields at varying distances from the focal plane of the DHM system are numerically reconstructed using various diffraction theories[4]. Digital in-line holographic microscopy (DIHM), originating from Gabor's holography[5], has been utilized to measure 3D positions and in-focus intensity maps of test samples over time[6]. Spatiotemporal tracking and label-free identification of various microparticles, including flow tracers[7,8], environmental pollutants[9-11], and biological specimens[12,13], can be quantitatively measured using DIHM. Meanwhile, one of the limitations of DIHM is the twin image problem caused by the loss of phase information during hologram recording[14].

Several instrumental and computational methods have been developed to recover the phase information for analyzing the 3D morphology of test samples. In off-axis DHM, the reference beam is tilted relative to the object beam to spatially separate the interference signals of real and twin images in the Fourier domain[15,16]. Phase-shifting digital holography was developed to reconstruct the phase information from multiple phase-shifted holograms by employing a phase-shifting device[17,18]. In DIHM, the iterative phase retrieval method was adopted to recover the phase map on the object plane based on the Gerchberg-Saxton algorithm[19-21] and multi-height phase recovery[22,23]. Since the recovered two-dimensional (2D) phase map is projected onto the image plane, 3D refractive index distribution can be reconstructed from the phase maps acquired by angle scanning[24]. Meanwhile, the instrumental approaches require more precise and complicated optical configurations compared to DIHM. If multiple intensity maps are required for reconstructing a phase map, it would be limited in the analysis of dynamic behaviors of moving samples. Additionally, it is challenging to reconstruct the full 3D morphology of a test sample from a single-shot hologram.

Recently, artificial intelligence (AI) has been widely applied to phase retrieval of DHM[25]. For dataset-driven approach, holographic datasets, consisting of the holograms and the corresponding in-focus amplitude and phase maps, are generated to train a neural network[26]. The trained neural network can predict the phase map of a test sample from its single-shot hologram without any additional imaging processing. However, the supervised learning-based model has a generalization problem by which the predictive performance is degraded for untrained input holograms.

For physics-driven approach, on the other hand, the basic principle of the Gerchberg-Saxton algorithm is implemented into neural networks[27,28]. The output complex field at the object plane is forward propagated to the input hologram plane, minimizing the loss between the input hologram and the forward-propagated intensity map. A self-supervised learning model based on physics consistency is trained with synthetic holograms to achieve better generalization performance in the hologram reconstruction process without the need for experimental datasets[29]. The physics-based neural network can reconstruct amplitude and phase maps at the object plane from a single-shot hologram without the requirement of hologram datasets.

However, the reconstruction of 3D morphology of an object from its single phase map still remains a challenge. To simulate the corresponding 3D complex light field from a single-shot hologram, a coordinate-based neural network, known as a neural field, can be employed for reconstructing the optical diffraction properties in 3D space[30]. Physical properties at spatiotemporal coordinates are predicted by the neural network and mapped to the sensor domain. The loss between the simulated and measured data is then minimized to optimize the neural network. Neural fields have been utilized for volume rendering at different viewing direction[31], computed tomography[32], 3D refractive index mapping based on intensity diffraction tomography[33], and computational adaptive optics in widefield microscopy[34]. By integrating the physics-driven and coordinate-based neural networks in order to model the diffraction theory, it would be possible to reconstruct the 3D complex light field and 3D morphology of an object.

In this study, a novel AI-based DIHM technique for single-shot reconstruction of 3D morphology, named MorpHoloNet, is proposed. MorpHoloNet is based on the physics-driven and coordinate-based neural networks to simulate the optical diffraction of coherent incident light propagating through a 3D phase shift distribution. The loss between the simulated and input holograms on the sensor plane is minimized to optimize the proposed MorpHoloNet. The reconstruction performance of 3D morphology using MorpHoloNet is evaluated by using synthetic holograms of ellipsoids and experimental holograms of various biological cells. Spatiotemporal variations in 3D location, orientation, and morphology of biological cells are successfully reconstructed from consecutive single-shot holograms. The proposed MorpHoloNet would be utilized to analyze spatiotemporal tracking of rotational behaviors and morphological deformations of biological cells under various microfluidic and biochemical conditions.

## 2. Results

### 2.1. MorpHoloNet workflow

Fig. 1 shows the workflow of MorpHoloNet for reconstructing the 3D morphology of microparticles from their single-shot holograms. The 3D space through which the incident laser beam propagates is expressed by a 3D Cartesian coordinate system (Fig. 1a). The intervals along the *x* and *y* axes are the magnified pixel lengths Δ*x* and Δ*y* of an image sensor, respectively. The interval along the *z* axis is Δ*z*. 3D coordinates (*x*, *y*, *z*) are fed into MorpHoloNet to predict the corresponding object value (*o*) for determining whether the finite voxel belongs to a medium (*o* = 0) or an object (*o* = 1). The main purpose of MorpHoloNet is to solve the inverse problem of reconstructing 3D complex light field and refractive index distribution in 3D space in front of the focal plane of an objective lens from a holographic image obtained by DIHM (Fig. 1b).

The training process of MorpHoloNet is divided into two steps. In the first step, MorpHoloNet is pre-trained with prior knowledge about the approximate 3D location (*x′*, *y′*, *z′*) of a target object, obtained by conventional hologram reconstruction and autofocusing algorithms (Fig. 1c). The angular spectrum method (ASM) for wave propagation is employed to acquire reconstructed images at different depths away from the original holographic image of the object[4,6]. Tamura of the gradient focus function is utilized to search the in-focus image with sharp edges and its corresponding depth (*z′*) from the reconstructed images[35]. In-plane (*x′*, *y′*) position is determined by finding the centroid of the object on the in-focus image. The 3D morphology reconstructed by MorpHoloNet is likely to be centered around the approximate location (*x′*, *y′*, *z′*). To incorporate this prior knowledge into the training process, a normalized Gaussian distribution centered at (*x′*, *y′*, *z′*) is used for pre-training of MorpHoloNet. The normalized Gaussian distribution (*f*) is defined as follows:

$$f(x,y,z) = \exp\left(-\frac{(x-x')^2}{2\sigma_x^2} - \frac{(y-y')^2}{2\sigma_y^2} - \frac{(z-z')^2}{2\sigma_z^2}\right) \quad (1)$$

where *σ<sub>x</sub>*, *σ<sub>y</sub>*, and *σ<sub>z</sub>* are the standard deviations along the *x*, *y*, and *z* axes, respectively. At each depth $z_i$, the difference between object value array $o_i$ and the corresponding *f*(*x*, *y*, $z_i$) is minimized by reflecting the prior knowledge during the pre-training process. This pre-training step prevents MorpHoloNet from getting stuck in local optima and aids to reach rapid convergence to a solution.

After pre-training, MorpHoloNet is trained based on the wave propagation principle and boundary conditions to reconstruct the 3D complex light field ($U$) and 3D morphology of the object from its hologram $H$ at $z_0 = 0$ ($L \times M$ pixels). The depth-wise length of the 3D space for training MorpHoloNet is set longer than $z'$. It is then discretized into $z_0, z_1, \ldots, z_N$ with intervals of $\Delta z$. The object values for boundary conditions ($o_{BC}$) are assumed to be 0 at boundaries of the 3D space ($x_{BC} = \Delta x, L\Delta x; y_{BC} = \Delta y, M\Delta y; z_{BC} = 0, z_N$) (Fig. 1d). In the 3D space, the propagation of the incident laser beam is simulated based on ASM and object array $o_i$ at $z_i$ ($i = 1, 2, \ldots, N$) (Fig. 1e). The initial value of the incident laser beam $U_N$ at $z_N$ is assumed to be $U_N = \bar{H}^{0.5}$. Since $U_N$ does not always equal to $\bar{H}^{0.5}$ for an arbitrary hologram, the tensor associated with $U_N$ is set as a trainable parameter. If the finite voxel is completely filled with the object ($o = 1$), the corresponding phase shift ($\phi$) is defined as follows:

$$\phi = \frac{2\pi(n_{\text{obj}} - n_{\text{med}})\Delta z}{\lambda} \quad (2)$$

where $n_{\text{obj}}$ and $n_{\text{med}}$ are the refractive indices of the object and the medium, respectively. Since the light absorption in biological cells is minimal, the imaginary part of $n_{\text{obj}}$ is assumed to be 0. $\lambda$ denotes the wavelength of the incident laser beam. The initial value of $\phi$ is roughly set based on the refractive index obtained from literature or empirical estimates. The tensor associated with $\phi$ is set as a trainable parameter to adaptively adjust during the training process.

The complex light field $U_i$ at $z_i$ with the phase shift induced by $o_i$ is calculated by multiplying $U_i$ by $\exp(j\phi o_i)$. The subsequent complex light field $U_{i-1}$ at $z_{i-1}$ is then simulated by using ASM as follows:

$$U_{i-1} = \text{ASM}(U_i \exp(j\phi o_i); \Delta z) \quad (3)$$

where ASM(·) represents the wave propagation equation of ASM (see Materials and Methods for details). By propagating the complex light field from $U_N$ to $U_0$, the tensors associated with the object arrays in the 3D space are connected by implementing ASM(·). Finally, the difference between the reconstructed hologram $|U_0|^2$ and the experimental hologram $H$ is minimized to optimize MorpHoloNet for predicting the 3D morphology of the object. After training MorpHoloNet, 3D refractive index distribution $n(x, y, z)$ can be calculated as follows:

$$n(x, y, z) = n_{\text{med}} + \frac{\lambda \phi^* o^*(x, y, z)}{2\pi \Delta z} \quad (4)$$

where $\phi^*$ and $o^*$ are the optimized phase shift and object array values after training. Since the values of $o^*$ are trained between 0 and 1, the refractive index variation can also be measured. Additionally, the 3D morphology can be visualized by adopting the marching cubes algorithm[36].

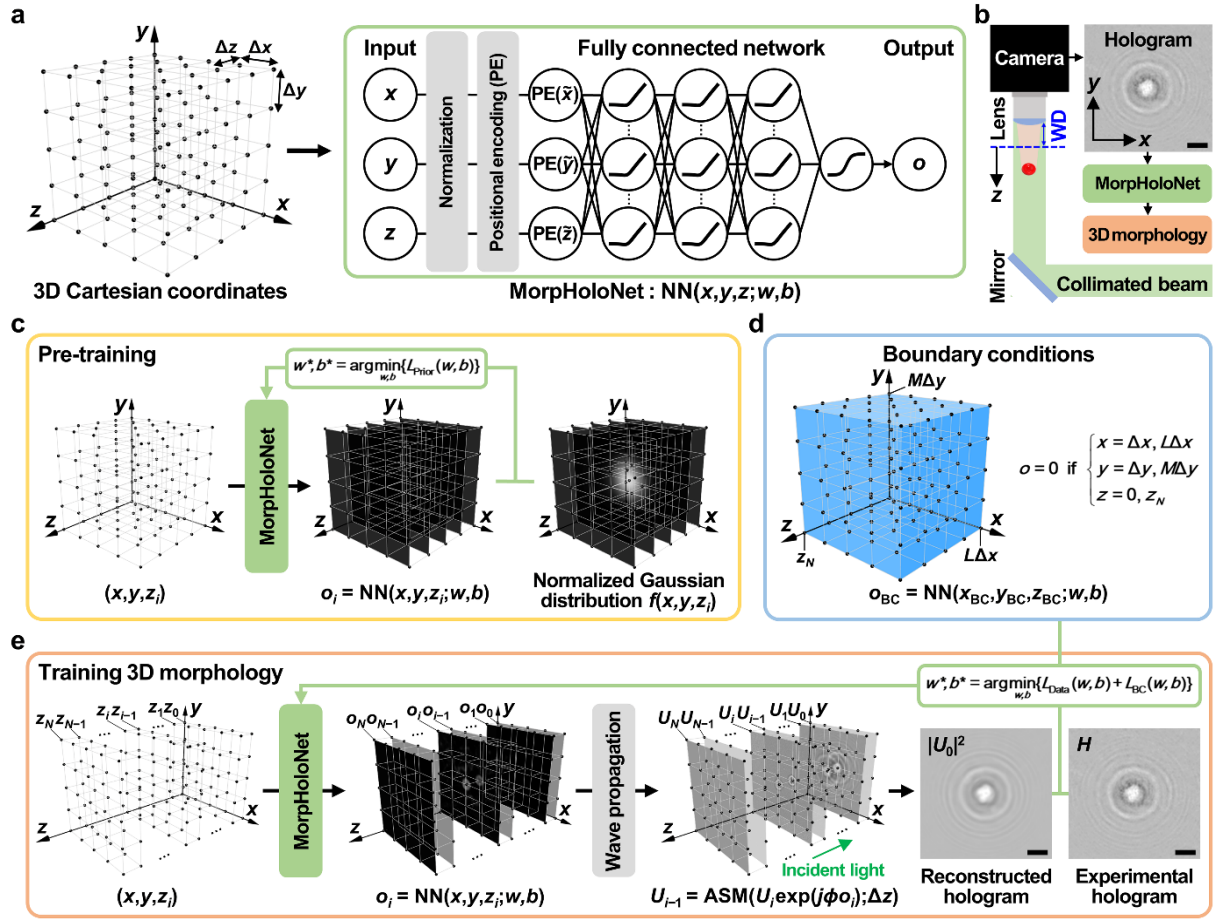

**Fig. 1**. Overall workflow of MorpHoloNet for single-shot reconstruction of the three-dimensional (3D) morphology of a microscale object. (a) Model architecture of MorpHoloNet. (b) Digital in-line holographic microscopy system for recording holograms of an object. (c) Pre-training of MorpHoloNet using prior knowledge about the approximate 3D location of the object. Training MorpHoloNet with (d) boundary conditions and (e) the angular spectrum method (ASM) for wave propagation. Scale bars are 10 μm.

## 2.2. Phase retrieval from synthetic hologram using MorpHoloNet

To verify the phase retrieval performance, MorpHoloNet is trained by using a synthetic hologram of three phase objects, as shown in Fig. 2. The phase objects $α$, $β$, and $γ$ induce phase shifts of $\pi/2$, $\pi/3$, and $\pi/6$ in the incident complex light field ($U_N$ = 0.707) at $z$ = 100 μm (Fig. 2b), respectively. The synthetic hologram is then simulated by propagating the phase-shifted wave over a depth of 100 μm by employing ASM. Fig. 2a shows the intensity maps reconstructed at depths of 25, 50, 75, and 100 μm using MorpHoloNet and ASM. The intensity maps reconstructed by conventional ASM exhibit the twin image problem caused by the loss of phase information at $z$ = 0 μm. However, MorpHoloNet eliminates diverging twin image signals. Fig. 2c is the phase map at $z$ = 100 μm reconstructed by MorpHoloNet. Fig. 2d shows the absolute errors between the ground truth and reconstructed phase maps. The overall 2D shape and phase information of the phase objects are accurately reconstructed with a small mean absolute error of 0.0018 rad. Some errors are observed in the concave regions of the phase objects.

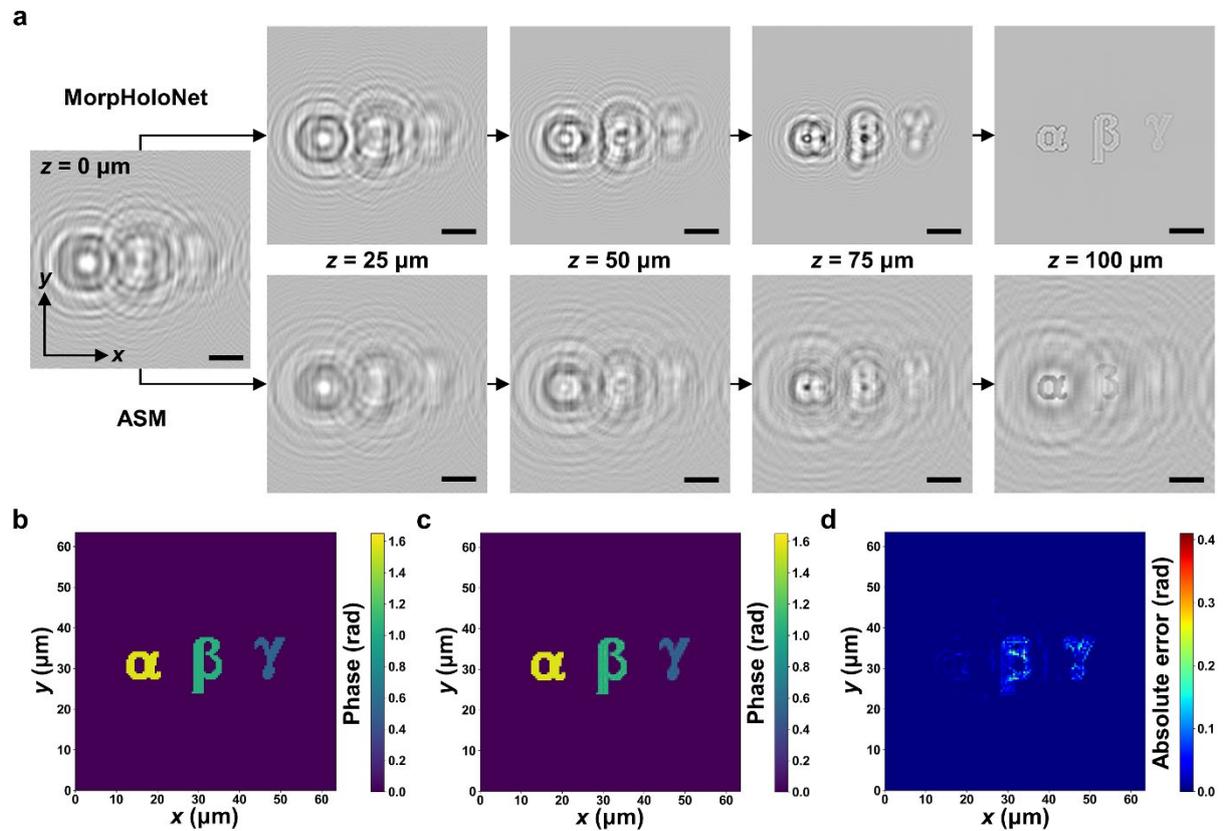

**Fig. 2.** Phase retrieval performance of MorpHoloNet. (a) Intensity maps at depths (*z*) of 25, 50, 75, and 100 μm are reconstructed from a synthetic hologram (*z* = 0 μm) of phase objects *α*, *β*, and *γ* by using MorpHoloNet and the angular spectrum method (ASM). (b) Ground-truth and (c) reconstructed phase maps of the phase objects. (d) Absolute errors between the ground-truth and reconstructed phase maps. Scale bars are 10 μm.

### 2.3. Reconstruction of 3D morphology from synthetic holograms using MorpHoloNet

To verify the performance of 3D morphology reconstruction, MorpHoloNet is trained by using synthetic holograms of an ellipsoid, as shown in Fig. 3. The semi-axis lengths of the ellipsoid along the *x*, *y*, and *z* axes are set as 2 μm, 2 μm, and 3 μm, respectively. The inclination angle ($\theta$) is defined as the angle between the *z*-axis and the major axis of the ellipsoid. The ellipsoid is located at *z* = 100 μm. The refractive indices of the ellipsoid and medium are set as 1.4 and 1.33, respectively. Synthetic holograms of the ellipsoid inclined with $\theta$ = 0°, 30°, 45°, 60°, and 90° are simulated by adopting discrete dipole approximation codes[37,38]. Fig. 3a shows the intensity maps of the ellipsoid with $\theta$ = 0°, reconstructed at depths of 25, 50, 75, and 100 μm using MorpHoloNet and ASM. Fig. 3b represents the intensity maps of the ellipsoid with different inclination angles at *z* = 0 μm, along with insets showing the corresponding reconstructed images at *z* = 100 μm using ASM.

The ground-truth and reconstructed morphologies of the ellipsoid are depicted in Figs. 3c and 3d, respectively. As a result, both 3D morphology and orientation of the ellipsoid are reconstructed from its single-shot holograms. For this, MorpHoloNet is trained for each ellipsoid hologram three times to evaluate the measurement error in single-shot reconstruction of 3D morphology. The reconstructed inclination angle and the lengths of semi-major and semi-minor axes are evaluated in terms of the average and standard deviation values, as summarized in Table 1. The root mean squared errors of the reconstructed inclination angle, semi-major axis, and semi-minor axis are 0.658 °, 0.085 μm, and 0.097 μm, respectively.

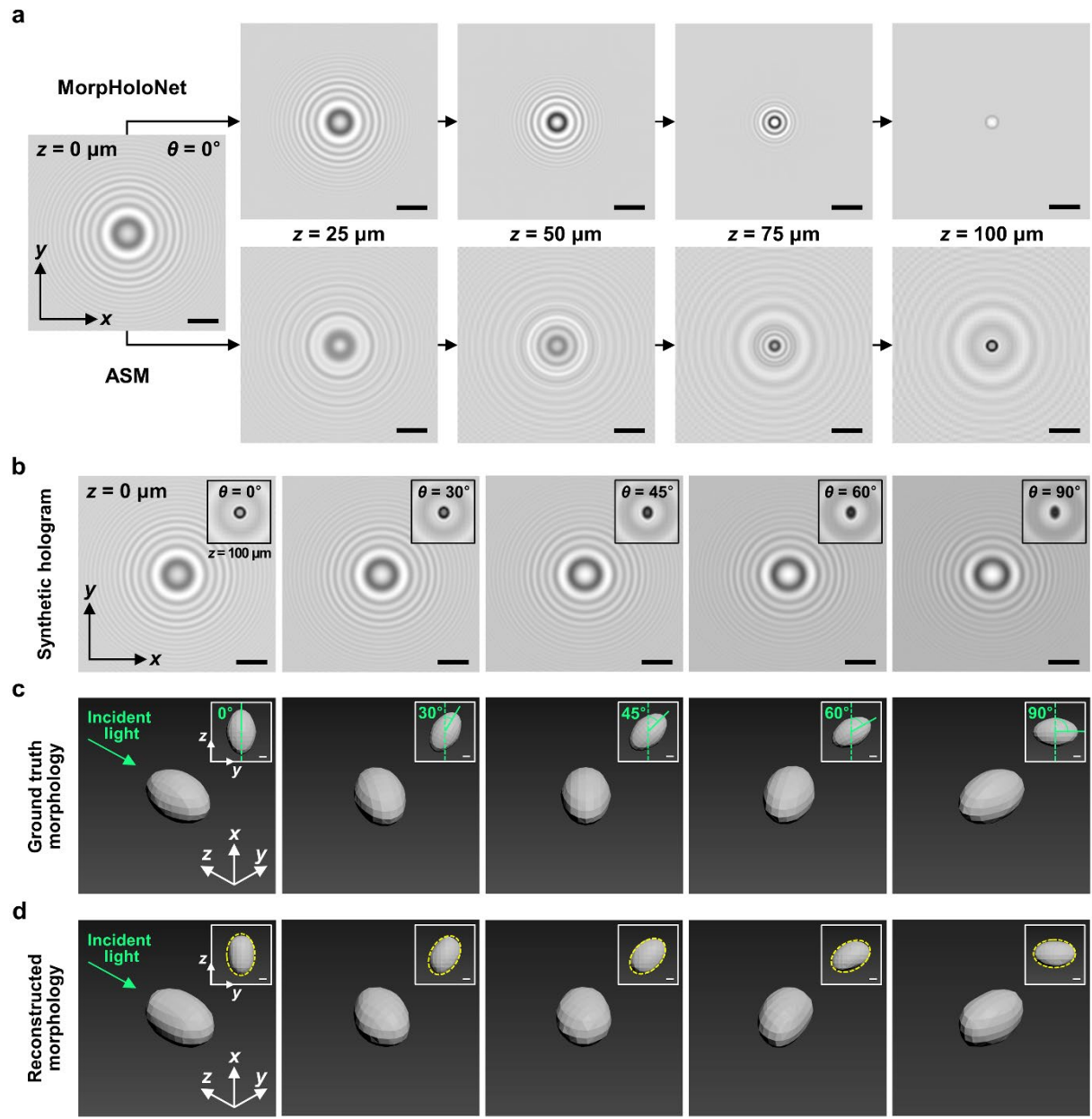

**Figure 3**. Single-shot reconstruction of three-dimensional morphology from synthetic holograms of an ellipsoid by using MorpHoloNet. (a) Intensity maps at depths ($z$) of 25, 50, 75, and 100 μm are reconstructed from a synthetic hologram ($z = 0$ μm) of the ellipsoid with an inclination angle ($\theta$) of 0° using MorpHoloNet and the angular spectrum method (ASM). (b) Synthetic holograms of the ellipsoid inclined with $\theta$ = 0°, 30°, 45°, 60°, and 90°. Intensity maps at $z$ = 100 μm reconstructed by ASM are inset in black boxes. (c) Ground truth and (d) reconstructed morphologies of the ellipsoid with different $\theta$. Top view images are inset in white boxes. Scale bars: (a), (b) 10 μm; (c), (d) 1 μm.

**Table 1.** Comparison between the ground truth inclination angles of an ellipsoid and the corresponding reconstructed inclination angle, semi-major axis, and semi-minor axis values predicted by MorpHoloNet.

| Ground truth inclination angle (°) | 0 | 30 | 45 | 60 | 90 |
|---|---|---|---|---|---|
| Reconstructed inclination angle (°) | 0 | 30.5 ± 2.5 | 44.5 ± 1.3 | 61.3 ± 2.5 | 90.3 ± 0.6 |
| Reconstructed semi-major axis (μm) | 3.07 ± 0.21 | 2.87 ± 0.02 | 2.91 ± 0.04 | 3.02 ± 0.07 | 3.07 ± 0.04 |
| Reconstructed semi-minor axis (μm) | 1.90 ± 0.04 | 1.97 ± 0.06 | 2.19 ± 0.03 | 2.00 ± 0.09 | 1.97 ± 0.11 |

**2.4. Reconstruction of 3D morphology of biological cells using MorpHoloNet**

To verify the performance of 3D morphology reconstruction for biological cells, MorpHoloNet is trained by using holographic images of several biological cells, as shown in Figs. 4 and 5. Fig. 4a shows typical optical microscope images of isotonic, hypotonic, and hypertonic red blood cells (RBCs) immersed in 6% w/w polyvinylpyrrolidone (PVP)-phosphate buffered saline (PBS) solution, distilled water, and 10% w/w sodium chloride (NaCl) solution, respectively. While isotonic RBCs preserve their biconcave shape, the membranes of hypotonic and hypertonic RBCs are deformed due to the osmotic pressure gradient between the intracellular contents of RBCs and the surrounding medium. The hypotonic RBCs swell to approximately twice the thickness of isotonic RBCs as solvent enters the cells. In contrast, the hypertonic RBCs have irregular and flattened membrane structures as solvent leaves the cells.

Fig. 4b represent the intensity maps of RBCs at $z = 0$ μm, along with the corresponding in-focus images reconstructed by ASM for comparison. For training MorpHoloNet, the initial refractive index ($n_{obj}$) of RBCs is set as 1.4[39]. The refractive indices ($n_{med}$) of 6% w/w PVP-PBS solution, distilled water, and 10% w/w NaCl solution are set as 1.344, 1.333, and 1.347, respectively. The 3D morphology of each RBC is reconstructed from its single-shot hologram with the aid of MorpHoloNet (Fig. 4c). The volume of the hypotonic RBC is measured to be 2.15 times larger than that of the isotonic RBC. The volume of the hypertonic RBC is measured to be 2.84 times smaller than the isotonic RBC. The reconstructed 3D morphologies of three different RBCs closely resemble the morphological features of RBCs observed in their optical microscope images.

Fig. 4d shows cross-sectional refractive index maps at depths near the centroids of RBCs. The refractive index of the isotonic RBC is relatively higher at the rim of thick edges, compared to the concave center region of the RBC. The dilution of intracellular contents owing to incoming water in the hypotonic RBC results in a lower refractive index, compared to the isotonic RBC. The refractive index of the hypotonic RBC at $z = 44$ μm ranges from 1.339 to 1.344. If the refractive index of the hypotonic RBC is assumed to be proportional to the concentration of intracellular contents, the measured refractive index is reasonable when considering the volume expansion ratio. The dehydration and shrinkage of the hypertonic RBC result in a higher refractive index, compared to the isotonic RBC.

To qualitatively evaluate the measurement accuracy of refractive index maps, phase maps of RBCs are acquired using the iterative phase retrieval method[19-21] (Fig. 4e). The refractive index and phase maps of each RBC exhibit structural similarity. While the hypotonic RBC becomes a weaker phase object after swelling, the phase shift of the hypertonic RBC largely increases after dehydration. The refractive index difference ($\Delta n = n_{obj} - n_{med}$) of the hypotonic RBC is less than half that of the isotonic RBC. In addition, the thickness of the hypotonic RBC is approximately twice that of the isotonic RBC. Therefore, if the intracellular contents of isotonic and hypotonic RBCs are preserved, there should be little change in the average phase shift values of the RBCs. However, the phase shift value of the hypotonic RBC is measured to be significantly lower than the value estimated based on the expansion ratio between the isotonic and hypotonic RBCs. These results imply that the proposed MorpHoloNet can predict both the variations in 3D cell morphology and intracellular refractive index of RBCs.

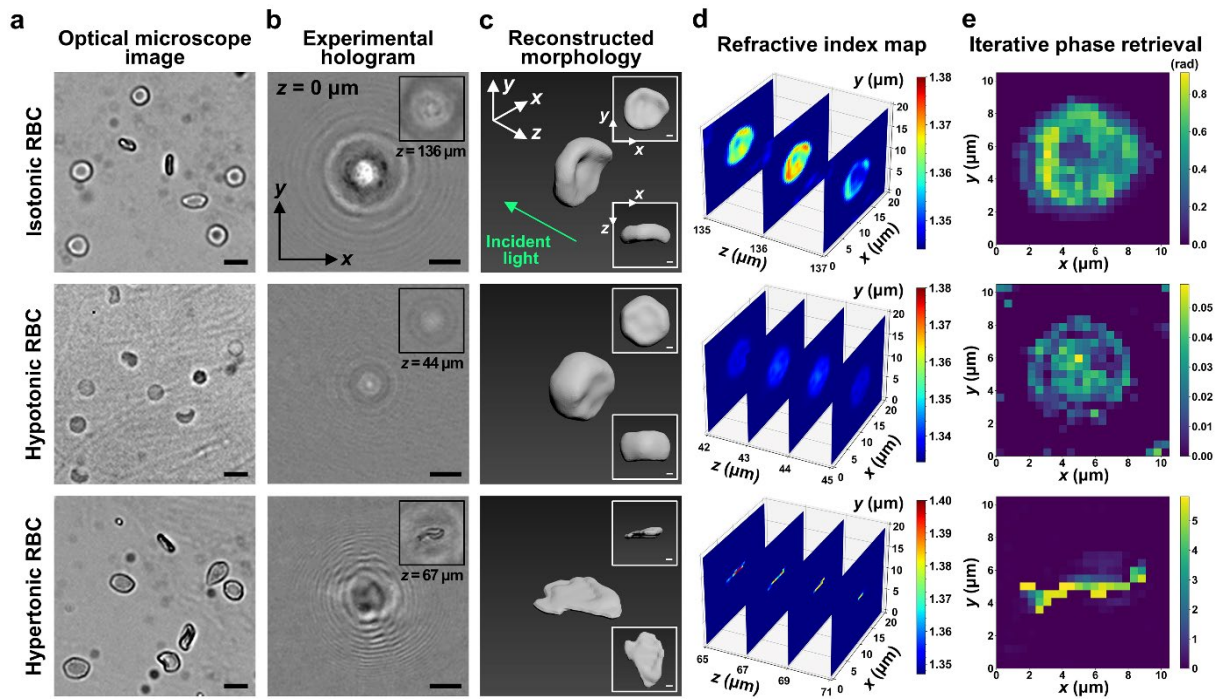

**Figure 4**. Single-shot reconstruction of three-dimensional (3D) morphology from experimental holograms of isotonic, hypotonic, and hypertonic red blood cells (RBCs) using MorpHoloNet. (a) Optical microscope images and (b) experimental holograms of RBCs. In-focus RBC images reconstructed using the angular spectrum method are inset in black boxes. (c) 3D morphologies

and (d) refractive index maps of RBCs reconstructed by MorpHoloNet. Front view and top view images are inset in white boxes. (e) Phase maps of RBCs reconstructed using the iterative phase retrieval method. Scale bars: (a), (b) 10 μm; (c) 1 μm.

Fig. 5a shows typical scanning electron microscope (SEM) images of unicellular microorganisms, including *Microcystis protocystis* and *Escherichia coli*. Fig. 5b presents the intensity maps of the microorganisms immersed in PBS at $z = 0$ μm, along with the corresponding in-focus images reconstructed by ASM. For training MorpHoloNet, the initial refractive index ($n_{obj}$) of *M. protocystis* and *E. coli* is set as $1.38^{40,41}$ and $1.39^{42}$, respectively. MorpHoloNet is found to successfully reconstruct the 3D morphologies and refractive index maps of the tested microorganisms from their single-shot holograms (Figs. 5c and 5d). The reconstructed 3D morphologies of *M. protocystis* and *E. coli* exhibit spherical and ellipsoidal shapes, respectively. Although the iterative phase retrieval method can provide phase maps containing information about the projected morphologies of microorganisms, it is difficult to reconstruct 3D morphological information from the phase maps (Fig. 5e). Therefore, MorpHoloNet can effectively obtain both 3D morphology and orientation of microorganisms from their single-shot holograms.

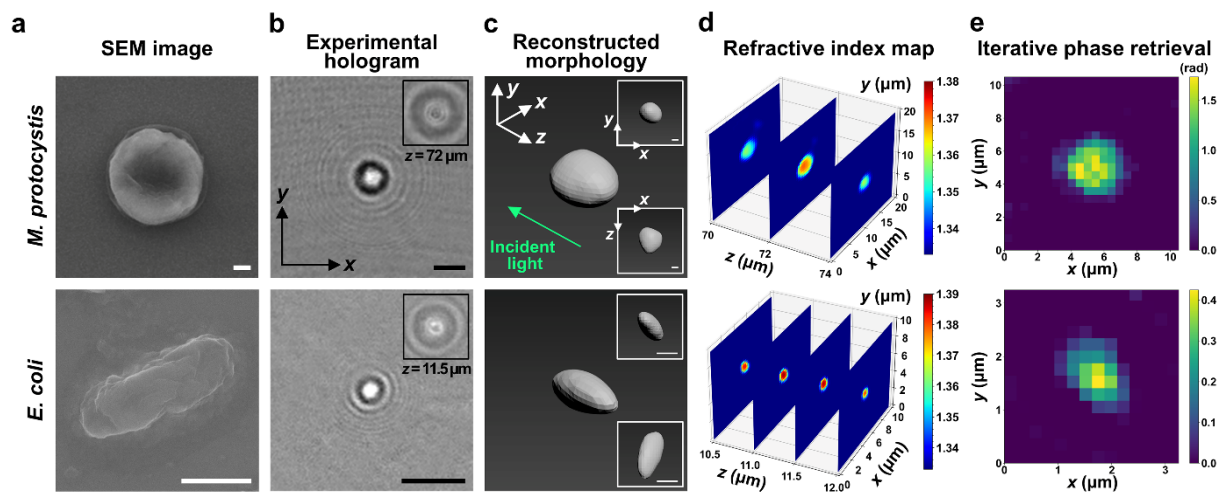

**Figure 5**. Single-shot reconstruction of three-dimensional (3D) morphology from experimental holograms of *Microcystis protocystis* and *Escherichia coli* using MorpHoloNet. (a) Scanning electron microscope (SEM) images and (b) experimental holograms of *M. protocystis* and *E. coli*. In-focus images of the microorganisms reconstructed using the angular spectrum method are inset in black boxes. (c) 3D morphologies and (d) refractive index maps of the microorganisms reconstructed by MorpHoloNet. Front view and top view images are inset in white boxes. (e) Phase maps of the microorganisms reconstructed using the iterative phase retrieval method. Scale bars: (a), (c) 1 μm; (b) 10 μm.

## 2.5. Spatiotemporal tracking of biological cells using MorpHoloNet

The single-shot reconstruction capability of MorpHoloNet enables reliable spatiotemporal tracking of biological cells from consecutive holograms, as shown in Figs. 6 and 7. Holograms of a tumbling RBC are consecutively acquired in a fully-developed Hagen-Poiseuille flow passing through a micropipe with an internal radius ($R$) of 150 μm (Fig. 6d). The working fluid is viscoelastic 6% w/w PVP-PBS solution. Fig. 6a shows a typical RBC hologram captured at $z = 0$ μm and time $t = 0$ s. Intensity maps of the RBC are reconstructed at depths of 30, 60, 90, and 115 μm using MorpHoloNet and ASM. Fig. 6b shows the intensity maps of the tumbling

RBC captured at $t$ = 0 s, 0.26 s, 0.51 s, and 0.76 s, along with the corresponding reconstructed images at $z$ = 115 μm by using ASM. Temporal variations of the 3D morphology of the tumbling RBC are reconstructed from consecutive holograms using MorpHoloNet (Figs. 6c and 6e). The 3D morphology can be robustly reconstructed even in the presence of overlapping holographic signals of adjacent RBCs.

RBCs moving in a shear flow exhibit various rotational behaviors and deformation characteristics depending on microfluidic conditions, including viscosity contrast and shear rate of the flow[43-45]. To interpret the observed rotational behavior of the RBC, the experimental conditions used in this study are compared with those reported in previous studies. The viscosities of RBC cytosol ($\mu_{cytosol}$) and blood plasma are assumed to be 0.006 Pa·s[46] and 0.0012 Pa·s[47], respectively. The viscosity of 6% w/w PVP-PBS solution ($\mu_{PVP}$) is measured to be 0.03 Pa·s. The viscosity contrast ($\mu_c = \mu_{cytosol}/\mu_{PVP}$), defined as the ratio of the viscosity of RBC cytosol to that of 6% w/w PVP-PBS solution, is approximately 0.2. The flow rate ($Q$) of the viscoelastic flow is set as 1 μl/min. The radial position ($r$) of the tumbling RBC is measured to be 71 μm. The corresponding shear rate ($\dot{\gamma}(r) = 4Qr/\pi R^4$) around the tumbling RBC is 2.98 s$^{-1}$.

Since the characteristic relaxation time of RBCs ($\tau$) is proportional to the dynamic viscosity of the surrounding medium, it is assumed to be 0.0016 s in blood plasma[45]. The characteristic time of RBCs in 6% w/w PVP-PBS solution is 25 times higher than that in blood plasma. The dimensionless shear rate ($\dot{\gamma}^* = \dot{\gamma}\tau$) is calculated as 0.12. RBCs flowing in a rectangular channel under the given dimensionless shear rate exhibited a tank-treading motion or a rolling motion of stomatocytes[43,45]. By dividing the dimensionless shear rate by the characteristic time of RBCs in PBS, RBCs under the shear rate of 100 s$^{-1}$ exhibited various shapes, including discocytes, normal stomatocytes, and deformed stomatocytes[44].

Meanwhile, the RBC observed in this study shows rotational behaviors which are similar to both tank-treading and tumbling stomatocytes. The rotational motion observed in this study is somewhat different from previous studies. This difference might be caused by different flow characteristics in cylindrical and rectangular microchannels. Simultaneous measurement of the 3D positional and morphological information of RBCs using MorpHoloNet would be a useful tool for analyzing the 3D rotating behaviors of RBCs under various microfluidic conditions.

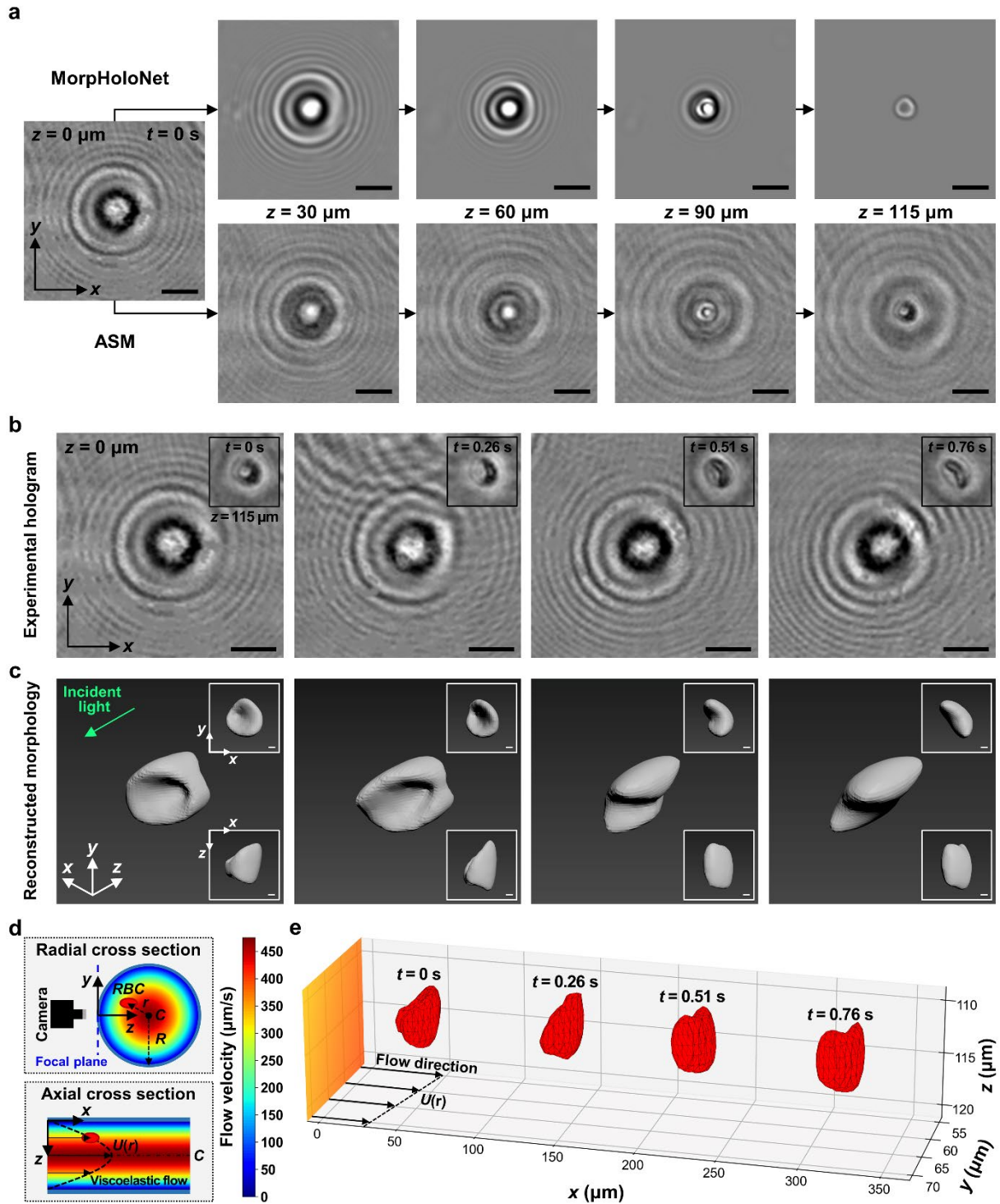

**Figure 6**. Single-shot reconstruction of three-dimensional (3D) morphology from consecutive holograms of a tumbling red blood cell (RBC) moving in a viscoelastic flow using MorpHoloNet. (a) Intensity maps at depths ($z$) of 30, 60, 90, and 115 μm are reconstructed

from an experimental hologram ($z$ = 0 μm) of the RBC captured at time $t$ = 0 s using MorpHoloNet and the angular spectrum method (ASM). (b) Consecutive holograms of the tumbling RBC captured at $t$ = 0 s, 0.26 s, 0.51 s, and 0.76 s. Intensity maps at $z$ = 115 μm reconstructed by ASM are inset in black boxes. (c) Temporal variation in 3D morphology of the tumbling RBC reconstructed using MorpHoloNet. Front view and top view images are inset in white boxes. (d) Radial and axial cross sections of the viscoelastic flow passing through a micropipe with an internal radius ($R$) of 150 μm. The tumbling RBC is located at a radial position ($r$) away from the center ($C$) of the micropipe. $U(r)$ denotes the velocity profile of the fully-developed Hagen-Poiseuille flow, where the maximum flow velocity is 472 μm/s. (e) Spatiotemporal tracking of rotational behavior and morphological deformation of the tumbling RBC. Scale bars: (a), (b) 10 μm; (c) 1 μm.

Fig. 7a shows a typical hologram of a free-swimming *E. coli* in PBS captured at $z$ = 0 μm and $t$ = 0.4 s. Intensity maps of the *E. coli* are reconstructed at depths of 5, 10, 15, and 20 μm using MorpHoloNet and ASM. Fig. 7b depicts the intensity maps of the *E. coli* captured at $t$ = 0 s, 0.2 s, 0.4 s, 0.6 s, and 0.8 s, along with the corresponding in-focus images reconstructed by ASM. Temporal variations of the 3D positional, orientational, and morphological information of the *E. coli* are reconstructed from consecutive holograms using MorpHoloNet (Figs. 7c–f). While the *E. coli* has a rod-shape in its SEM image (Fig. 5a), the reconstructed morphology is somewhat elongated along the z-axis. Since the propagating distance Δ$z$ is similar to the thickness of *E. coli*[48], this difference is presumed to be caused by the uncertainty encountered in depth localization of the *E. coli* during the AI training process. The swimming speed of the *E. coli* is measured as 14.2 ± 1.4 μm/s, which is similar to the average speed of *E. coli* ranging from 14 to 19 μm/s in previous studies[48,49]. It is difficult to observe the flagella of

*E. coli* due to the limited spatial resolution of the DIHM system used in this study. Since the swimming direction of the *E. coli* does not align with its orientation, it is presumed to be in a tumbling motion. These experimental results imply that the proposed MorpHoloNet can usefully utilized to extract the spatiotemporal variations in 3D positional, orientational, and morphological information of biological cells from consecutive single-shot holograms.

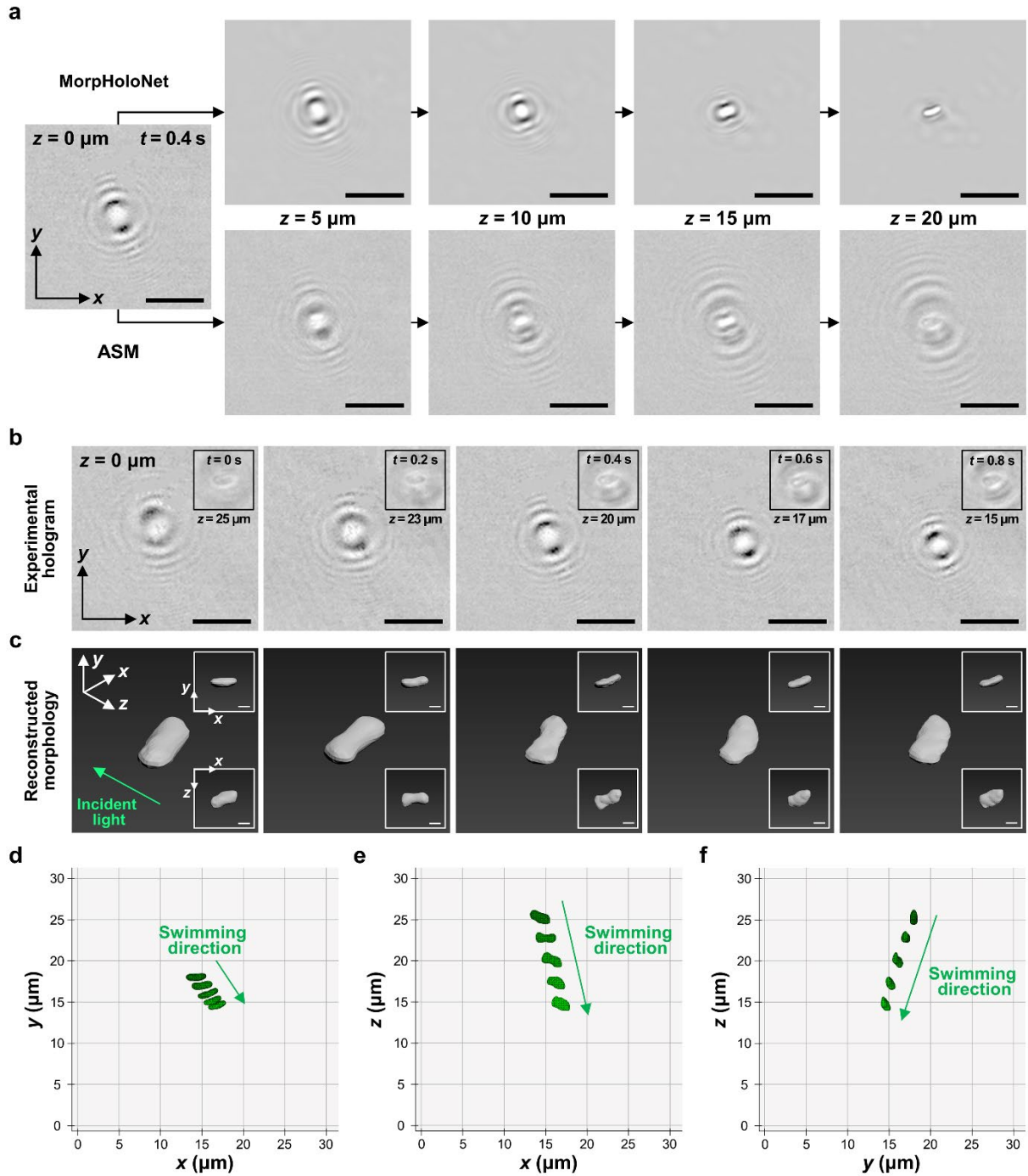

**Figure 7**. Single-shot reconstruction of three-dimensional (3D) morphology from consecutive holograms of a free-swimming *Escherichia coli* using MorpHoloNet. (a) Intensity maps at depths (*z*) of 5, 10, 15, and 20 μm are reconstructed from an experimental hologram (*z* = 0 μm) of the *E. coli* captured at time *t* = 0 s using MorpHoloNet and the angular spectrum method

(ASM). (b) Consecutive holograms of the *E. coli* captured at $t$ = 0 s, 0.2 s, 0.4 s, 0.6 s, and 0.8 s. In-focus images of the *E. coli* reconstructed using ASM are inset in black boxes. (c) Temporal variation in 3D morphology of the *E. coli* reconstructed using MorpHoloNet. Front view and top view images are inset in white boxes. Spatiotemporal tracking of rotational behavior and morphological deformation of the *E. coli* projected onto the (d) *xy*, (e) *xz*, and (f) *yz* planes. Scale bars: (a), (b) 10 μm; (c) 1 μm.

## 3. Discussion

MorpHoloNet has several key advantages compared to the existing phase retrieval methods. It can reconstruct the 3D morphology and refractive index distribution of a biological cell from its single-shot hologram without the need for angle scanning for acquiring multiple tomographic images. Compared to the iterative phase retrieval method[19-21], the phase information not only on the object plane but also in 3D space can be reconstructed from a single-shot hologram captured by DIHM. The intensity maps reconstructed by MorpHoloNet exhibit the performance of denoising and twin image removal (Figs. 6 and 7). Additionally, the 3D centroid of a test sample can be automatically searched from its initially estimated location used for pre-training process. Compared to the dataset-driven approach[25], MorpHoloNet does not suffer from the generalization problem of supervised learning. Therefore, the translational and rotational dynamic behaviors of deforming biological cells can be simultaneously analyzed from consecutive single-shot holograms.

On the other hand, further in-depth research is required to improve the performance of MorpHoloNet. For instance, the convergence problem of a physics-driven neural network needs to be further stabilized. The pre-training of MorpHoloNet is a key step to make it converge on the desirable location and morphology. It would become a more practical approach if MorpHoloNet could converge without the pre-training step for estimating the approximate 3D location of a test sample. The depth-wise elongation in the reconstructed morphology, observed in cases of the ellipsoid with $\theta = 45°$ (Fig. 3d) and *E. coli* (Fig. 5a), needs to be resolved by reducing the uncertainty encountered in depth localization at discretized coordinates. If the imaginary part of $n_{obj}$ is not negligible, the initial assumption about the incident laser beam ($U_N = \bar{H}^{0.5}$) can induce reconstruction errors. If $n_{med}$ could also be learned as a trainable parameter, it would be more convenient to reconstruct 3D morphology without prior knowledge of the medium. The excessive number of epochs can lead to overfitting, caused by the remaining noise in a hologram. Memory allocation should also be optimized to extend the field-of-view and depth-wise light propagating distance. In this study, the training time takes approximately 30 minutes. It may increase as the input image size, the number of epochs, and $N$ increase. Especially, it should be reduced to rapidly process consecutive holograms for spatiotemporal cell tracking. These further improvement and optimization would enhance the throughput and measurement accuracy, making MorpHoloNet a more practical and efficient tool.

The proposed MorpHoloNet has promising potential for use in a wide range of biomedical and engineering applications. Temporal variations in morphology and refractive index of biological cells according to extracellular conditions can be quantitatively measured using MorpHoloNet (Fig. 4). The morphological shapes of hypotonic and hypertonic RBCs are similar to those of spherocytes and echinocytes, respectively. It implies that MorpHoloNet can be utilized to measure the 3D morphologies and dynamic behaviors of various types of RBCs with hematologic diseases, such sickle cell anemia, thalassemia, and malaria infection. Unlike the existing supervised learning-based DIHM method that needs to be retrained with new hologram datasets in order to measure the orientational information of new types of RBCs[12],

MorpHoloNet can reconstruct the comprehensive 3D morphological information from a single-shot hologram without acquiring hologram datasets. MorpHoloNet would lead to deeper understanding of microbial behaviors, such as cell-surface interactions during bacterial adhesion and biofilm formation[13]. Beyond biological cells, it can also be utilized to analyze the rotational behaviors of other particulate objects with microstructures within a sufficiently large field of view in real time.

In summary, a new AI-based DIHM technique is developed for single-shot reconstruction of 3D morphology of biological cells by integrating the physics-driven and coordinate-based neural networks. The performance of 3D morphology reconstruction using MorpHoloNet is systematically demonstrated by using single-shot synthetic and experimental holograms. MorpHoloNet would be utilized to investigate the translational and rotational dynamic behaviors and morphological changes of biological cells under various microenvironmental conditions.

## 4. Materials and methods

### 4.1. Sample preparation

All experiments associated with RBCs were conducted based on the principles of the Declaration of Helsinki and approved by the institutional review board of Pohang University of Science and Technology (POSTECH IRB, PIRB-2024-A001). The blood sample collected from a healthy donor was mixed with different types of media, including PBS (Gibco™, USA) solution of 6% w/w PVP ($M_w$ = 360 kDa, Sigma-Aldrich, USA), distilled water, and 10% w/w NaCl solution, to a hematocrit of 0.5%. For acquiring consecutive RBC holograms in a viscoelastic flow, RBCs suspended in 6% w/w PVP-PBS solution were injected into a perfluoroalkoxy (PFA, $R$ = 150 μm, As One, Japan) microtube using a syringe pump (KD

Scientific, USA). The refractive index, dynamic viscosity, and osmolarity of 6% w/w PVP-PBS solution are 1.344, 0.03 Pa·s, and 290 mOsm, respectively. The PFA tube and an objective lens were immersed in 6% w/w PVP-PBS solution to suppress optical distortions. An optical microscope (Eclipse 80i, Nikon, Japan) was utilized to observe morphological differences between isotonic, hypotonic, and hypertonic RBCs.

*M. protocystis* was collected from freshwater of Yangju Wondang reservoir. Sterile BG11 broth (MB-B0872, KisanBio, Korea) was used for incubation of *M. protocystis*. An experimental chamber was kept at 23 ± 2°C and 45 ± 5% humidity. The photoperiod inside the chamber was set to 16 hours of light and 8 hours of darkness, with a light intensity of 25 µmol/m$^2$/s. *M. protocystis* was diluted by using Jirisan mineral water (Ourhome, Korea). It was then aliquoted into 1 ml vials and stored at 4 °C. For SEM imaging, 10 µl of aliquoted *M. protocystis* was dropped on a silicon wafer chip (1 × 1 cm$^2$) and dried at room temperature. After sputter coating the sample with platinum, a field emission SEM (FE-SEM, JSM-7100F, JEOL, Japan) was utilized to acquire SEM images of the dried *M. protocystis*.

*E. coli* was purchased from the American Type Culture Collection (ATCC, USA). Agar plates dehydrated with yeast extract and tryptone were streaked with *E. coli*. After overnight incubation at 37 °C, the plates were stored at 4 °C. To make *E. coli* broth, one colony was taken from the streak plate and inoculated into 5 ml of sterile Luria broth (Sigma-Aldrich, USA). *E. coli* was cultured at 37 °C for 24 hours, while being shaken at 75 rpm. After centrifugation at 1800 rpm for 3 min, the remaining *E. coli* pellet was washed with PBS. It was centrifuged again to remove PBS supernatant. The washed *E. coli* was diluted in 20 ml of PBS. It was then aliquoted into 10 ml and stored at 4 °C. For SEM imaging, the washed *E. coli* was fixed in 10% formalin solution (Sigma-Aldrich, USA) for 30 minutes at room temperature. After centrifugation and washing, the fixed *E. coli* was dehydrated in a graded ethanol series up to 100%. 10 µl of diluted *E. coli* was placed onto a silicon wafer chip and dried at room temperature. After platinum coating, SEM images of the dried *E. coli* were captured by using the FE-SEM.

**4.2. Hologram recording**

Holographic images of biological cells were acquired by using a DIHM system (Fig. 1b). A collimated beam was generated by using a green laser source ($\lambda$ = 532 nm, 70 mW, USA), a spatial filter, and a convex lens. A mirror was used to align the collimated beam perpendicular to the detector plane. A high-speed CMOS camera (pixel size = 10 µm, UX100, Photron, Japan) was utilized for capturing holographic interference signals of biological cells with a resolution of 1280 × 1024 pixels. A 20x water-immersion objective lens (numerical aperture NA = 0.5, working distance WD = 2 mm. Nikon, Japan) was attached for acquiring holographic images of RBCs and *M. protocystis*. The magnified pixel size and spatial resolution ($\Delta = 0.61\lambda/NA$) of the DIHM system with a 20x objective lens were 0.5 µm and 649 nm, respectively. A 40x water-immersion objective lens (NA = 0.8, WD = 2 mm. Nikon, Japan) was utilized for acquiring holographic images of *E. coli*. The magnified pixel size and spatial resolution of the DIHM system with a 40x objective lens were 0.25 µm and 406 nm, respectively. The ensemble-averaging method was adopted to remove background noises of raw holograms.

**4.3. Numerical wave propagation using ASM**

The numerical propagation of the complex light field (*E*) was simulated by using ASM[6]. The complex wave field propagated along the *z*-axis with an interval of $\Delta z$ can be expressed as follows:

$$E(\xi, \eta; z + \Delta z) = \mathrm{ASM}(E(x, y; z); \Delta z)$$

$$= F^{-1}\left[F(E(x,y;z))\exp\left(ik\Delta z\sqrt{1-\left(\frac{\lambda x}{L\Delta x}\right)^2 - \left(\frac{\lambda y}{M\Delta y}\right)^2}\right)\right] \quad (5)$$

where $\xi$ and $\eta$ are the spatial coordinates of the propagated wave field $E$. $x$ and $y$ are the spatial coordinates on the rectangular grid of input $E$ with a resolution of $L \times M$ pixels. $\Delta x$ and $\Delta y$ represent the magnified pixel lengths in the $x$ and $y$ directions, respectively. $k$ denotes the wave number in a medium, defined as $k = 2\pi n_{\mathrm{med}}/\lambda$. $F$ and $F^{-1}$ represent the fast Fourier transform and inverse fast Fourier transform, respectively.

### 4.4. Architecture and implementation of MorpHoloNet

MorpHoloNet is based on the fully connected neural network NN($x, y, z; w, b$) consisting of one input layer, one Fourier feature projection layer for positional encoding[50], three dense layers, and one output layer (Fig. 1a). $w$ and $b$ represent the weights and biases within the neural network. Between the input and Fourier feature projection layers, the in-plane coordinates ($x$, $y$) are normalized by $L\Delta x$ and $M\Delta y$, respectively. The depth-wise coordinate $z$ is normalized by the maximum value of $z_N$. The normalized 3D coordinates $(\tilde{x}, \tilde{y}, \tilde{z})$ ranges between 0 and 1. The output size of the Fourier feature projection and three dense layers is set as 128. The activation functions of the dense and output layers are swish and sigmoid functions, respectively. For pre-training, the mean squared error between the object value array $o_i$ and the corresponding normalized Gaussian distribution $f(x, y, z_i)$ at each depth $z_i$ is minimized by employing the Adam optimizer (Fig. 1c). The training loss from prior knowledge ($L_{\mathrm{Prior}}$) about the approximate 3D location of a target object is defined as follows:

$$L_{\mathrm{Prior}}(w, b) = \frac{\sum_{l=1}^{L}\sum_{m=1}^{M}\sum_{i=0}^{N}(\mathrm{NN}(x_l, y_m, z_i; w, b) - f(x_l, y_m, z_i))^2}{LM(N+1)} \quad (6)$$

where $x_l$ and $y_m$ are the discrete $x$ and $y$ positions on 2D complex light field at each depth $z_i$.

After pre-training, the wave propagation principle of ASM and boundary conditions are utilized to train MorpHoloNet. The training losses from the boundary conditions on the planes perpendicular to $x$-axis ($L_{BC,x}$), $y$-axis ($L_{BC,y}$), and $z$-axis ($L_{BC,z}$) are defined as follows:

$$L_{BC,x}(w,b) = \frac{\sum_{l\in\{1,L\}}^{L}\sum_{m=1}^{M}\sum_{i=0}^{N}(NN(x_l, y_m, z_i; w, b))^2}{2M(N+1)} \quad (7)$$

$$L_{BC,y}(w,b) = \frac{\sum_{l=1}^{L}\sum_{m\in\{1,M\}}^{M}\sum_{i=0}^{N}(NN(x_l, y_m, z_i; w, b))^2}{2L(N+1)} \quad (8)$$

$$L_{BC,z}(w,b) = \frac{\sum_{l=1}^{L}\sum_{m=1}^{M}\sum_{i\in\{0,N\}}(NN(x_l, y_m, z_i; w, b))^2}{2LM} \quad (9)$$

where the total training loss of the boundary conditions ($L_{BC}$) is the summation of $L_{BC,x}$, $L_{BC,y}$, and $L_{BC,z}$ (Fig. 1d). The training loss of the wave propagation principle is defined as follows:

$$L_{Data}(w,b) = \frac{\sum_{l=1}^{L}\sum_{m=1}^{M}(|U_0(x_l, y_m)|^2 - H(x_l, y_m))^2}{LM} \quad (10)$$

where $|U_0|^2$ and $H$ are the simulated and experimentally obtained intensity maps at $z = 0$ μm. Typical size of $H$ is 128 × 128 pixels. For training 3D morphology, the summation of $L_{Data}$ and $L_{BC}$ is optimized by using the Adam optimizer. The $\Delta z$ value for both synthetic holograms and experimental holograms of RBCs and *M. protocystis* is set to 1 μm, while that of experimental holograms of *E. coli* is set to 0.5 μm. The epoch and learning rate for pre-training are 300 and 0.001. The epoch and learning rate for training 3D morphology are 700 and 0.0001. The epoch and learning rate values can be adjusted through fine-tuning.

### 4.5. Development environment

MorpHoloNet was trained using Python 3.6.7, Anaconda3-4.5.11, PyCharm (JetBrains, Czech Republic), TensorFlow-gpu 2.4.1, NVIDIA CUDA toolkit 11.0, and cuDNN 8.2.1. A desktop computer used in this study is composed of Nvidia GeForce RTX 3090 GPU, AMD Ryzen 5950X CPU, and 128 GB RAM. MATLAB R2021a software was used for digital image processing of holograms. Synthetic holograms of ellipsoids were simulated by using the Python packages of HoloPy 3.5.0 and discrete dipole approximation codes[37,38]. The Python package of PyMCubes 0.1.4 was used to convert object value arrays into a digital asset exchange (DAE) file. Autodesk 3ds Max 2025 software was utilized for rendering the DAE file. COMSOL Multiphysics 5.6 software was used to simulate the velocity profile of a fully developed Hagen-Poiseuille flow.

**References**

1   Kim, M. K. Principles and techniques of digital holographic microscopy. *SPIE Rev.* **1**, 018005 (2010).

2   Osten, W. *et al.* Recent advances in digital holography. *Appl. Opt.* **53**, G44-G63 (2014).

3   Javidi, B. *et al.* Roadmap on digital holography. *Opt. Express* **29**, 35078-35118 (2021).

4   Goodman, J. W. *Introduction to Fourier optics*. (Roberts and Company publishers, 2005).

5   Gabor, D. A new microscopic principle. *Nature* **161**, 777-778 (1948).

6   Choi, Y. S., Seo, K. W., Sohn, M. H. & Lee, S. J. Advances in digital holographic micro-PTV for analyzing microscale flows. *Opt. Lasers Eng.* **50**, 39-45 (2012).

7   Katz, J. & Sheng, J. Applications of holography in fluid mechanics and particle dynamics. *Annu. Rev. Fluid Mech.* **42**, 531-555 (2010).

8   Ling, H. *et al.* High-resolution velocity measurement in the inner part of turbulent boundary layers over super-hydrophobic surfaces. *J. Fluid Mech.* **801**, 670-703 (2016).

9   Wu, Y. C. *et al.* Air quality monitoring using mobile microscopy and machine learning. *Light Sci. Appl.* **6**, e17046 (2017).

10  Go, T., Kim, J. & Lee, S. J. Three-dimensional volumetric monitoring of settling particulate matters on a leaf using digital in-line holographic microscopy. *J. Hazard. Mater.* **404**, 124116 (2021).

11  Kim, J., Kim, J., Kim, Y., Go, T. & Lee, S. J. Accelerated settling velocity of airborne particulate matter on hairy plant leaves. *J. Environ. Manage.* **332**, 117313 (2023).

12  Kim, Y., Kim, J., Seo, E. & Lee, S. J. AI-based analysis of 3D position and orientation of red blood cells using a digital in-line holographic microscopy. *Biosens. Bioelectron.* **229**, 115232 (2023).

13  Kim, J. & Lee, S. J. Digital in-line holographic microscopy for label-free identification and tracking of biological cells. *Mil. Med. Res.* **11**, 38 (2024).

14  Stoykova, E., Kang, H. & Park, J. Twin-image problem in digital holography - A survey. *Chin. Opt. Lett.* **12**, 060013 (2014).

15  Leith, E. N. & Upatnieks, J. Reconstructed wavefronts and communication theory. *JOSA* **52**, 1123-1130 (1962).

16  Park, Y. K., Depeursinge, C. & Popescu, G. Quantitative phase imaging in biomedicine.

*Nat. Photonics* **12**, 578-589 (2018).

17  Chang, M., Hu, C. P., Lam, P. & Wyant, J. C. High precision deformation measurement by digital phase shifting holographic interferometry. *Appl. Opt.* **24**, 3780-3783 (1985).

18  Awatsuji, Y. *et al.* Parallel two-step phase-shifting digital holography. *Appl. Opt.* **47**, D183-D189 (2008).

19  Gerchberg, R. W. A practical algorithm for the determination of plane from image and diffraction pictures. *Optik* **35**, 237-246 (1972).

20  Latychevskaia, T. & Fink, H. W. Solution to the twin image problem in holography. *Phys. Rev. Lett.* **98**, 233901 (2007).

21  Latychevskaia, T. Iterative phase retrieval for digital holography: Tutorial. *JOSA A* **36**, D31-D40 (2019).

22  Greenbaum, A. & Ozcan, A. Maskless imaging of dense samples using pixel super-resolution based multi-height lensfree on-chip microscopy. *Opt. Express* **20**, 3129-3143 (2012).

23  Rivenson, Y. *et al.* Sparsity-based multi-height phase recovery in holographic microscopy. *Sci. Rep.* **6**, 37862 (2016).

24  Kim, G. *et al.* Holotomography. *Nat. Rev. Methods Primers* **4**, 51 (2024).

25  Wang, K. *et al.* On the use of deep learning for phase recovery. *Light Sci. Appl.* **13**, 4 (2024).

26  Rivenson, Y., Zhang, Y., Günaydın, H., Teng, D. & Ozcan, A. Phase recovery and holographic image reconstruction using deep learning in neural networks. *Light Sci Appl.* **7**, 17141 (2018).

27  Wang, F. *et al.* Phase imaging with an untrained neural network. *Light Sci. Appl.* **9**, 77 (2020).

28  Zhang, F. *et al.* Physics-based iterative projection complex neural network for phase retrieval in lensless microscopy imaging. In *Proceedings of the IEEE/CVF Conference on Computer Vision and Pattern Recognition.* 10523-10531 (IEEE, 2021).

29  Huang, L., Chen, H., Liu, T. & Ozcan, A. Self-supervised learning of hologram reconstruction using physics consistency. *Nat. Mach. Intell.* **5**, 895-907 (2023).

30  Xie, Y. *et al.* Neural fields in visual computing and beyond. *Comput. Graph. Forum* **41**, 641-676 (2022).

31      Mildenhall, B. *et al.* Nerf: Representing scenes as neural radiance fields for view synthesis. *Commun. ACM* **65**, 99-106 (2021).

32      Sun, Y., Liu, J., Xie, M., Wohlberg, B. & Kamilov, U. S. Coil: Coordinate-based internal learning for tomographic imaging. *IEEE Trans. Comput. Imaging* **7**, 1400-1412 (2021).

33      Liu, R., Sun, Y., Zhu, J., Tian, L. & Kamilov, U. S. Recovery of continuous 3D refractive index maps from discrete intensity-only measurements using neural fields. *Nat. Mach. Intell.* **4**, 781-791 (2022).

34      Kang, I., Zhang, Q., Yu, S. X. & Ji, N. Coordinate-based neural representations for computational adaptive optics in widefield microscopy. *Nat. Mach. Intell.* **6**, 714-725 (2024).

35      Zhang, Y., Wang, H., Wu, Y., Tamamitsu, M. & Ozcan, A. Edge sparsity criterion for robust holographic autofocusing. *Opt. Lett.* **42**, 3824-3827 (2017).

36      Lorensen, W. E. Marching cubes: A high resolution 3D surface construction algorithm. *Comput. Graph.* **21**, 7-12 (1987).

37      Yurkin, M. A. & Hoekstra, A. G. The discrete-dipole-approximation code ADDA: Capabilities and known limitations. *J. Quant. Spectrosc. Radiat. Transf.* **112**, 2234-2247 (2011).

38      Wang, A. *et al.* Using the discrete dipole approximation and holographic microscopy to measure rotational dynamics of non-spherical colloidal particles. *J. Quant. Spectrosc. Radiat. Transf.* **146**, 499-509 (2014).

39      Mazigo, E. *et al.* Ring stage classification of *Babesia microti* and *Plasmodium falciparum* using optical diffraction 3D tomographic technique. *Parasit. Vectors* **15**, 434 (2022).

40      Gautam, R. *et al.* Nonlinear optical response and self-trapping of light in biological suspensions. *Adv. Phys.: X* **5**, 1778526 (2020).

41      Zhai, S. *et al.* Optical backscattering and linear polarization properties of the colony forming cyanobacterium *Microcystis*. *Opt. Express* **28**, 37149-37166 (2020).

42      Liu, P. Y. *et al.* Cell refractive index for cell biology and disease diagnosis: Past, present and future. *Lab Chip* **16**, 634-644 (2016).

43      Dupire, J., Socol, M. & Viallat, A. Full dynamics of a red blood cell in shear flow. *PNAS* **109**, 20808-20813 (2012).

44	Lanotte, L. *et al.* Red cells' dynamic morphologies govern blood shear thinning under microcirculatory flow conditions. *PNAS* **113**, 13289-13294 (2016).

45	Mauer, J. *et al.* Flow-induced transitions of red blood cell shapes under shear. *Phys. Rev. Lett.* **121**, 118103 (2018).

46	Wells, R. & Schmid-Schönbein, H. Red cell deformation and fluidity of concentrated cell suspensions. *J. Appl. Physiol.* **27**, 213-217 (1969).

47	Késmárky, G., Kenyeres, P., Rábai, M. & Tóth, K. Plasma viscosity: A forgotten variable. *Clin. Hemorheol. Microcirc.* **39**, 243-246 (2008).

48	Molaei, M., Barry, M., Stocker, R. & Sheng, J. Failed escape: Solid surfaces prevent tumbling of *Escherichia coli*. *Phys. Rev. Lett.* **113**, 068103 (2014).

49	Cheong, F. C. *et al.* Rapid, high-throughput tracking of bacterial motility in 3D via phase-contrast holographic video microscopy. *Biophys. J.* **108**, 1248-1256 (2015).

50	Tancik, M. *et al.* Fourier features let networks learn high frequency functions in low dimensional domains. *Adv. Neural inf. Process. Syst.* **33**, 7537-7547 (2020).